\documentclass[journal]{IEEEtran}

\usepackage{amsmath,amssymb,amsfonts}
\usepackage{algorithm}
\usepackage{graphicx}
\usepackage{textcomp}
\usepackage{booktabs}
\usepackage{multirow}
\usepackage{xcolor}
\usepackage{tikz}
\usepackage{hyperref}
\usepackage{subcaption}
\usepackage{bm}
\usepackage{array}
\usepackage{cite}
\usepackage{balance}
\usepackage{lettrine}
\usepackage[font=small]{caption}
\usepackage{algpseudocode}
\algrenewcommand\algorithmicrequire{\textbf{Input:}}
\algrenewcommand\algorithmicensure{\textbf{Output:}}

\newcommand{\eg}{\textit{e.g.}}
\newcommand{\etal}{\textit{et al.}}

\newcommand{\RR}{\mathbb{R}}
\newcommand{\ade}{\text{ADE}}
\newcommand{\fde}{\text{FDE}}
\newcommand{\softmax}{\mathrm{softmax}}

\graphicspath{{figures/}}

\begin{document}

\title{Social Graph Mamba: Forecasting Pedestrian Movements Based on Social Context}

\author{Hong-Son Nguyen, Yen-Chen Liu
 \thanks{This work was supported in part by the National Science and Technology Council (NSTC), Taiwan, under Grant NSTC 114-2218-E-006-021 and NSTC 114-2628-E-006-010. This research was also supported in part by the Higher Education Sprout Project, Ministry of Education to the Headquarters of University Advancement at National Cheng Kung University (NCKU).}
 \thanks{H.-S. Nguyen, and Y.-C. Liu are with the Department of Mechanical Engineering, National Cheng Kung University, Tainan 70101, Taiwan. Email: \{{\tt hongsonnguyen.haui@gmail.com, yliu@mail.ncku.edu.tw}\}.}%
}

\maketitle

\begin{abstract}
Forecasting pedestrian motion has always been fundamental for autonomous navigation in crowded environments. While attention-based methods achieve strong performance, they suffer from quadratic computational complexity in modeling social interactions, limiting scalability. Additionally, the existing methods often achieve high accuracy on prediction benchmarks at the individual level, but fail to fully capture the natural movement behaviors of crowds in real-world scenarios, particularly group structures. In this study, we propose \textbf{Social Graph Mamba} (SGM), a novel architecture that replaces attention-based social reasoning with Selective State Space Models (SSMs) operating on dynamically constructed interaction graphs. SGM introduces a dynamic interaction graph with social triplet factorization to decompose crowd interactions sequentially, and a community-aware module to effectively discover group structures via differentiable MinCut optimization and conditions both the embedding space and multi-modal decoder on group membership. Our experiments on standard benchmarks (ETH/UCY, SDD) demonstrate competitive performance with linear sequence complexity compared to quadratic attention-based methods. We further validate SGM in physical robot experiments by integrating predicted trajectories into a Social Force Model (SFM) for real-world implementation.
\end{abstract}


\section{Introduction}\label{sec:introduction}
Predicting the future trajectories of pedestrians is a primary capability for mobile robots, autonomous vehicles, and intelligent surveillance systems. In crowded environments, pedestrians do not move independently; their paths are shaped by rich social interactions---collision avoidance, group cohesion, lane formation, and goal-directed navigation through dense crowds~\cite{helbing1995social,alahi2016social}. Accurately modeling these interactions is therefore critical for safe and efficient autonomous navigation.

Early methods relied on hand-crafted models such as the SFM~\cite{helbing1995social}, which treats pedestrians as particles subject to attractive and repulsive forces. While physically interpretable, SFM struggles to capture the complexity and diversity of real human navigation behavior. The deep learning era brought data-driven approaches that significantly improved prediction accuracy. Social LSTM~\cite{alahi2016social} introduced social pooling to share hidden states between neighboring agents, while Social GAN~\cite{gupta2018social} further advanced multimodal prediction through generative adversarial training. Subsequent works introduced attention mechanisms~\cite{vemula2018social}, graph neural networks~\cite{mohamed2020social, shi2021sgcn}, and Transformer-based architectures~\cite{yu2020spatio, lee2024mart} that treat the crowd as a fully connected graph and learn pairwise interaction weights.

Despite their success, attention-based models uniformly suffer from $O(N^2)$ complexity in the number of agents $N$, both in computing attention weights and in message passing~\cite{vaswani2017attention,tang2022quadtree}. This quadratic scaling becomes a practical bottleneck in dense crowd scenarios with dozens or hundreds of pedestrians. Moreover, most existing models adopt an ego-centric view that prioritizes nearby agents based on spatial distance, neglecting the graph-theoretic structure of social interactions~\cite{kong2022gsta}. 

Recently, State Space Models (SSMs), particularly the Mamba architecture~\cite{gu2023mamba}, have emerged as a compelling alternative to Transformers. Mamba processes sequences in $O(N)$ time with input-dependent gating, making it naturally suited for modeling variable-length agent sequences. However, applying Mamba to social trajectory prediction requires addressing a fundamental challenge: SSMs are inherently sequential and process a single ordered sequence, whereas social interactions are unordered and graph-structured.

In this paper, we propose SGM, a novel architecture that bridges this gap through the following contributions:

\begin{itemize}
    \item An attention-free, graph-based architecture for prediction in social contexts, where sequential information is modeled through its backbone SSMs, achieving a balance between accuracy and computational efficiency.

    \item The model integrates social interaction features through a community-aware mechanism, enabling the discovery of groups without explicit group annotations, and conditioning both the contrastive embedding loss and the multimodal trajectory decoder on inferred group membership.


    \item We bridge the gap between trajectory prediction and practical robot navigation by designing a prediction framework that not only achieves accurate pedestrian forecasting but also satisfies the computational constraints of real-time deployment on mobile robotic platforms.
\end{itemize}

The remainder of this paper is organized as follows. Section~\ref{sec:related_works} reviews related work. Section~\ref{sec:proposed_method} presents the proposed SGM architecture in detail. Section~\ref{sec:experiments} describes the experimental setup and evaluation metrics. Section~\ref{sec:results_discussion} analyzes the benchmark comparisons, physical experiment results, and ablation study. Section~\ref{sec:conclusion} provides the conclusion and outlines potential directions for the paper.


\begin{figure*}[t]
    \vspace{0.5mm}
    \centering
    \centerline{\includegraphics[width=\textwidth]{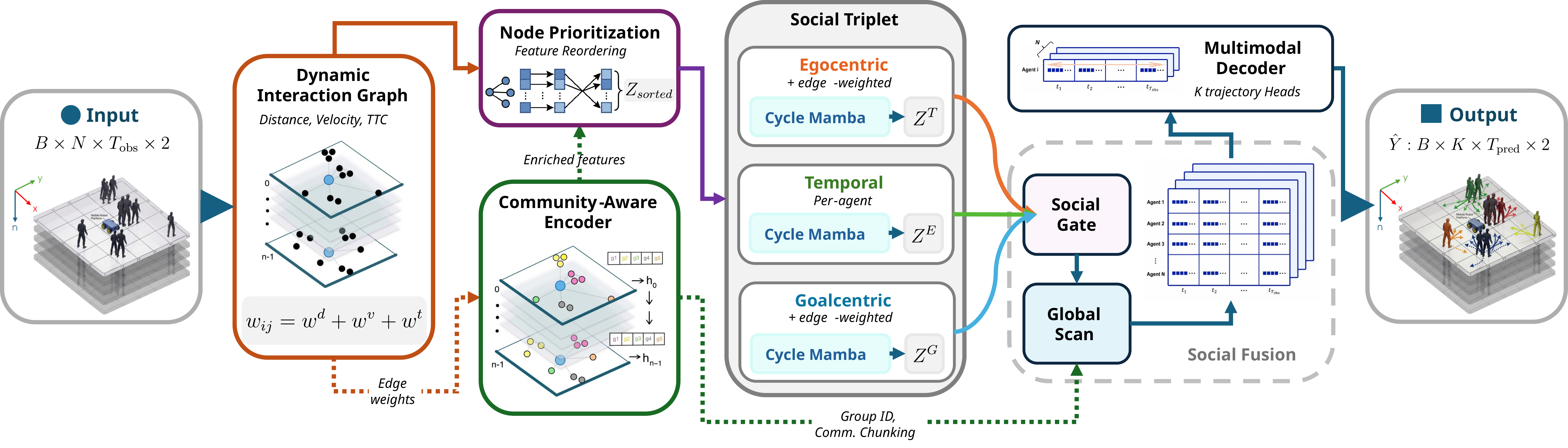}}
    \caption{The overview of Social Graph Mamba framework.}
    \label{fig:architecture}
\end{figure*}

\section{Related Works}\label{sec:related_works}

\subsection{Pedestrian Trajectory Prediction}

The SFM~\cite{helbing1995social} is the foundational physics-based approach, modeling pedestrians as particles governed by attractive forces toward goals and repulsive forces from obstacles and other pedestrians. Notable extensions include the Headed SFM~\cite{farina2017walking}, which accounts for pedestrian heading direction, while alternative physics-inspired approaches such as the Reciprocal Velocity Obstacle (RVO) ~\cite{van2008reciprocal} enforce reciprocal collision avoidance. While physically grounded, these models rely on hand-tuned parameters and fail to capture the diversity and context-dependence of real pedestrian behavior.


In response to these limitations, the deep learning approaches have become prevalent paradigms. The seminal Social LSTM~\cite{alahi2016social} introduced a social pooling mechanism to share hidden states among spatially proximate agents. Social GAN~\cite{gupta2018social} extended this with adversarial training and a variety pooling module for multimodal prediction. Social Attention~\cite{vemula2018social} replaced pooling with soft attention over neighboring agents.

Graph-based methods formalize social interactions as graph neural networks. Social-STGCNN~\cite{mohamed2020social} uses spatio-temporal graph convolutions on a distance-based adjacency matrix. SGCN~\cite{shi2021sgcn} introduces sparse graph convolution networks that distinguish between attractive and repulsive interactions. PECNet~\cite{mangalam2020not} conditions predictions on estimated endpoints. Trajectron++~\cite{salzmann2020trajectron++} combines graph attention with conditional variational autoencoders for heterogeneous agent types.

Transformer-based approaches~\cite{yu2020spatio, lee2024mart, yuan2021agentformer} treat the crowd as a set of tokens and use self-attention for both temporal and social reasoning. AgentFormer~\cite{yuan2021agentformer} jointly models temporal and social dimensions in a single Transformer. While powerful, all attention-based methods incur $O(N^2)$ complexity per layer.

\subsection{Group-Aware Trajectory Prediction}

Several works recognize that pedestrians often move in social groups. Social-BIGAT~\cite{kosaraju2019social} uses bicycle-GAN with graph attention to capture group dynamics. GP-Graph~\cite{bae2022gpgraph} explicitly models group-level trajectories alongside individual predictions. EvolveGraph~\cite{li2020evolvegraph} and MART~\cite{lee2024mart} dynamically evolve interaction graphs to capture changing group structures. However, most group-aware methods require ground-truth group annotations during training, which are expensive to obtain. Our community-aware module discovers groups in an unsupervised manner through differentiable optimization.

\subsection{State Space Models and Mamba}

Mamba has been successfully applied to language modeling~\cite{gu2023mamba}, vision~\cite{zhu2024vision}, and graph learning~\cite{wang2024graph}. The advanced Structured State Space Models (S4)~\cite{gu2022efficiently} demonstrated that linear recurrences with carefully parameterized state matrices can match or exceed Transformer performance on long-range sequence tasks with $O(N)$ complexity. Mamba~\cite{gu2023mamba} introduced the Selective SSM mechanism with input-dependent gating:
\begin{align}
    \bm{h}_t &= \bar{\bm{A}}_t \bm{h}_{t-1} + \bar{\bm{B}}_t \bm{x}_t \label{eq:ssm_state}\\
    \bm{y}_t &= \bm{C}_t \bm{h}_t \label{eq:ssm_output}
\end{align}
where $\bar{\bm{A}}_t$ and $\bar{\bm{B}}_t$ are input-dependent discretized parameters, enabling content-aware filtering.

By leveraging the strengths of SSMs in sequential processing with linear-time complexity and optimized memory consumption, Wang~\etal~\cite{wang2024graph} proposed \emph{Graph-Mamba}, a novel framework that utilizes SSMs for efficient long-range context modeling as an alternative to graph attention mechanisms. Graph-Mamba introduced two key ideas: ~\emph{node prioritization}, which converts an unordered graph into a meaningful sequence by sorting nodes according to centrality measures, and ~\emph{permutation-based training}, which injects Gaussian noise into centrality scores during training to encourage ordering invariance. These ideas provide a principled solution to the sequence-ordering problem inherent in applying SSMs to graphs. STG-Mamba~\cite{li2024stg} further extends Mamba to spatio-temporal graph forecasting for traffic prediction.

To reason over social interactions in pedestrian trajectory forecasting, Social-Mamba~\cite{luan2026socialmamba} leverages Mamba by decomposing crowd dynamics into temporal, ego-centric, and goal-centric SSM branches. Our Social Triplet Factorization (Sec.~\ref{sec:module4}) adopts this perspective-based decomposition as its conceptual foundation.

Our work builds upon the node prioritization and permutation training paradigm of Graph-Mamba~\cite{wang2024graph} and adapts it to the domain of pedestrian trajectory prediction with several key improvements: (i)~we replace static graph adjacency with dynamic, physics-aware interaction graphs whose edge weights encode proximity, velocity alignment, and time-to-collision; (ii)~we introduce a graph-aware social triplet factorization with edge-weighted token conditioning, direction-aware goal prediction, and learnable directional balance; (iii)~we propose adaptive Cycle Mamba for advanced directional balance via a single unidirectional scan; and (iv)~we integrate community-aware centrality into the prioritization scheme.

\section{Proposed Method}\label{sec:proposed_method}

\subsection{Problem Definition}

Trajectory prediction is performed in a robot-centric coordinate frame. Given agent positions $\bm{p}_i^t$, observations are transformed relative to the robot's last observed position $\bm{p}_0^{T_{\text{obs}}}$ as $\bar{\bm{p}}_i^t=\bm{p}_i^t-\bm{p}_0^{T_{\text{obs}}}$ The model takes $\{\bar{\bm{p}}_i^t\}_{t=1}^{T_{\text{obs}}}$ as input and predicts future trajectories of the surrounding pedestrians over a horizon of $T_{\text{pred}}$ steps. To account for motion uncertainty, it outputs $K$ trajectory hypotheses and their corresponding probabilities.

\subsection{Overview}

Our proposed SGM consists of six integrated modules, as illustrated in Fig.~\ref{fig:architecture}. The input observed trajectories are first processed by the dynamic interaction graph module, which computes physics-aware edge weights and spatial feature embeddings. The community-aware encoder augments graph centrality with community-level features, enabling information chunking and conditioned decoding in downstream processing. The topology-guided node prioritization module then sorts agents by graph centrality for principled SSM ordering. The sorted features are processed by the social triplet factorization, which decomposes interactions into temporal, egocentric, and goalcentric branches, each using Cycle Mamba. The dynamic social gate fuses the three branches, and the global interaction scan aggregates population-level context with information chunking along the agent dimension. Finally, the multimodal conditioned-decoder generates $K$ trajectory hypotheses.

\subsection{Dynamic Interaction Graph}\label{sec:inter_graph}

We construct a dynamic interaction graph $\mathcal{G} = (\mathcal{V}, \mathcal{E})$ where each node $v_i \in \mathcal{V}$ represents an agent in a graph and edge weight $w_{ij} \in \mathcal{E}$ encodes interaction strength. Unlike prior work that uses only distance-based adjacency~\cite{mohamed2020social}, our edge weights decompose into three physics-motivated components:
\begin{equation}\label{eq:edge_weight}
    w_{ij} = w_{ij}^{\text{dist}} + w_{ij}^{\text{vel}} + w_{ij}^{\text{ttc}}
\end{equation}

where, \textit{distance component} $w_{ij}^{\text{dist}} = \alpha\exp(-\|\Delta\bm{p}\|^2/\sigma_p)$ is a Gaussian kernel captures spatial proximity with $\alpha = 1.0$ is the distance weight and $\sigma_p = 2.0$ is the kernel bandwidth; \textit{velocity component} $w_{ij}^{\text{vel}} = \beta\max(0,\cos(\bm{v}_i,\bm{v}_j))$ is cosine similarity between velocity vectors captures motion coherence with $\beta = 0.5$ and the $\max(0, \cdot)$ ensures only aligned velocities contribute positively; and \textit{Time-to-Collision (TTC) component} $w_{ij}^{\text{ttc}} = \gamma/(\text{TTC}_{ij}+\varepsilon)$ is the inverse TTC captures imminent collision risk with $\text{TTC}_{ij} = \|\bm{p}_i - \bm{p}_j\| / \|\bm{v}_i - \bm{v}_j\|$ and $\gamma = 2.0$ if two agents are considered approaching, otherwise got 0. TTC weights are capped at 10.0 for numerical stability.

To ensure computational efficiency, we apply a distance threshold $d_{\max} = 10\text{m}$ and retain only the top-$k$ neighbors per agent ($k = 20$):
\begin{equation}
    \mathcal{E} = \{(i,j) : \|\bm{p}_i - \bm{p}_j\| < d_{\max},\; j \in \text{top-}k(w_{i\cdot})\}
\end{equation}

Raw position-velocity features $[\bm{p}_i^t, \bm{v}_i^t] \in \RR^4$ at each timestep are embedded through a 3-layer MLP into a $d_e$-dimensional space, producing node features $\bm{Z}^{(0)} \in \RR^{B \times N \times T_{\text{obs}} \times d_e}$.

\subsection{Topology-Guided Node Prioritization}\label{sec:module23}

Applying SSMs to graph-structured data requires converting the unordered node set into a sorted sequence. We adopt Graph-Mamba's \emph{node prioritization} and \emph{permutation-based training} framework. Whereas Graph-Mamba operates on static graphs with fixed adjacency, our prioritization uses the \emph{dynamic}, physics-aware edge weights $w_{ij}$ from~\eqref{eq:edge_weight}. This means centrality scores change at every timestep and reflect real-time social interaction strength rather than static topology. Specifically, we use weighted-degree centrality $H_i = \sum_j w_{ij}$, where $w_{ij}$ encodes proximity, velocity alignment, and collision risk (Sec.~\ref{sec:inter_graph}). Agents are sorted in \emph{ascending} order so that the most influential agents occupy later positions, where the Mamba hidden state has accumulated the richest context.

When the community-aware module is active (Sec.~\ref{sec:community}), we further augment the centrality with community importance via~\eqref{eq:comm_centrality}, allowing group-bridge nodes and large-community members to receive higher priority.

After sorting, an input projection maps features from $d_e$ to the hidden dimension $d$, producing $\bm{Z}^{(1)} \in \RR^{B \times N \times T_{\text{obs}} \times d}$.

\begin{algorithm}[!htbp]
\caption{Graph-Aware Social Triplet Factorization}
\label{alg:triplet}
\begin{algorithmic}[1]
\Require Sorted features $\bm{Z}^{(1)} \in \RR^{B \times N \times T \times d}$, ego index $e$, sorted edge weights $\bm{W}$
\Ensure Branch outputs $\bm{Z}^T, \bm{Z}^E, \bm{Z}^G$
\State \textit{\textbf{Temporal branch} (per-agent, no cross-agent info)}:
\State $\bm{Z}^T \leftarrow \text{CycleMamba}^{(2)}(\text{reshape}(\bm{Z}^{(1)},\; BN{\times}T{\times}d))$
\State \textit{\textbf{Egocentric branch} (edge-weighted token)}:
\State $\bm{z}_{ego} \leftarrow \text{MLP}(\bm{Z}_e^{(1)}[:, -1, :])$ \Comment{Ego token from last obs.}
\State $\tilde{\bm{z}}_{ego}^{(j)} \leftarrow w_{ej} \cdot \bm{z}_{ego} \quad \forall j$ \Comment{Scale by edge weight (Eq.~\ref{eq:edge_weight})}
\State $\bm{Z}'^E_j \leftarrow [\tilde{\bm{z}}_{ego}^{(j)};\; \bm{Z}_j^{(1)}] \quad \forall j$
\State $\bm{Z}^E \leftarrow \text{CycleMamba}^{(2)}(\bm{Z}'^E)$
\State \textit{\textbf{Goalcentric branch} (direction-aware, edge-weighted)}:
\State $\bm{z}_{goal} \leftarrow f_{\text{goal}}(\text{flatten}(\bm{Z}_e^{(1)}[:, -k:, :]))$ \Comment{$k$ steps}
\State $\tilde{\bm{z}}_{goal}^{(j)} \leftarrow w_{ej} \cdot \bm{z}_{goal} \quad \forall j$ \Comment{Scale by edge weight (Eq.~\ref{eq:edge_weight})}
\State $\bm{Z}'^G_j \leftarrow [\tilde{\bm{z}}_{goal}^{(j)};\; \bm{Z}_j^{(1)}] \quad \forall j$
\State $\bm{Z}^G \leftarrow \text{CycleMamba}^{(2)}(\bm{Z}'^G)$
\State return $\bm{Z}^T,\; \bm{Z}^E,\; \bm{Z}^G$
\end{algorithmic}
\end{algorithm}

\subsection{Social Triplet Factorization}\label{sec:module4}

Relative to Social-Mamba's social triplet, our design introduces three principled enhancements that close a fundamental graph-awareness gap: \emph{edge-weighted token conditioning}---the injected ego and goal tokens are multiplied by the sorted edge weight $w_{ij}$ derived from Sec.~\ref{sec:inter_graph}, so each neighbor receives a conditioning signal whose strength is directly proportional to its dynamic interaction intensity with the focal agent; a \emph{direction-aware goal predictor} that concatenates the last $k{=}3$ observed feature vectors of the ego agent (instead of only the final state) before projecting to the goal token, encoding travel direction and velocity; and \emph{learnable directional balance} in Cycle Mamba, replacing the fixed equal-weight average of forward and backward outputs with a learned scalar $\theta$.

Given input $\bm{x}=[\bm{x}_1, \ldots, \bm{x}_T]$ and $\bm{x}_\text{bwd} = [\bm{x}_T, \ldots, \bm{x}_1]$, single Mamba scan on $\bm{x}_{\text{cycle}} = [\bm{x}_\text{bwd}, \bm{x}] \in \RR^{2T \times d}$ produces $\bm{y}_{\text{cycle}}$, which is split into backward and forward components:
\begin{align}
    \bm{y} &= \sigma(\theta)\,\bm{y}_{\text{fwd}} + (1 - \sigma(\theta))\,\bm{y}_{\text{bwd}} \label{eq:cycle_combine}
\end{align}
where $\bm{y}_\text{fwd} = \bm{y}_{\text{cycle}}[T\!+\!1:2T]$, $\bm{y}_\text{bwd} = \text{flip}(\bm{y}_{\text{cycle}}[1:T])$, and $\theta \in \RR$ is a learnable scalar parameter initialized to $0$, yielding a balanced combination at the start of training ($\sigma(0) = 0.5$). The model adapts $\theta$ over training; trajectory prediction typically benefits from recency, so $\theta$ tends to converge above $0.5$.

Crucially, the forward scan is initialized with the hidden state accumulated from the backward sequence, and vice versa, creating deep cross-directional context propagation with only half the parameters of a traditional bidirectional model. The complete social triplet is given in Algorithm~\ref{alg:triplet}.

Unlike Social-Mamba~\cite{luan2026socialmamba}, which only predicts ego-agent trajectory, ours jointly forecasts for all $P$ pedestrians simultaneously, naively running the egocentric and goalcentric branches $P$ times would be prohibitive. 

\subsection{Community-Aware Encoding}\label{sec:community}

We introduce a community-aware module that discovers group structure in an unsupervised manner, contributes feature enrichment, and conditions the final multimodal decoder accordingly.

\subsubsection{Community Feature Enrichment}

Given node features $\bm{Z} \in \RR^{B \times N \times d}$ and adjacency matrix $\bm{A} \in \RR^{B \times N \times N}$, we compute soft community assignments $\bm{S} = \softmax\left(\text{MLP}([\bm{Z}\;;\; \bm{d}])\right) \in \RR^{B \times N \times C}$, where $\bm{d}_i$ is the node degree, concatenated with node features as input, and $C$ is the maximum number of communities (an upper bound that adapts to the actual scene). When $N < C$, excess community logits are masked with $-\infty$ before softmax. After that, the community information is injected into node features through learnable community prototypes $\bm{P} \in \RR^{C \times d}$. The enriched features $\bm{Z}'$ replace the original features before entering the Social Triplet Factorization. Subsequently, the members of a tight group share community-conditioned mode preferences, and the community-aware centrality
\begin{equation}\label{eq:comm_centrality}
    H_i^{\text{comm}} = \sum_c S_{ic} \cdot \frac{|\mathcal{C}_c|}{N} + 0.5 \cdot \frac{\mathcal{H}(\bm{s}_i)}{\log C},
\end{equation}
which augments graph centrality during node prioritization (Sec.~\ref{sec:module23}); high-entropy assignments act as bridge scores.  The Community Feature Enrichment process is described in Algorithm~\ref{alg:community}.

\subsubsection{MinCut Optimization}

We employ a MinCut loss--the soft, differentiable relaxation of the classical normalized-cut objective~\cite{shi2000normalized}, introduced as a graph-pooling layer by Bianchi~\etal~\cite{bianchi2020spectral} and repurposed here for soft community discovery. that maximizes intra-community edge density without balance constraints:
\begin{equation}\label{eq:mincut}
    \mathcal{L}_{\text{cut}} = -\frac{\text{Tr}(\bm{S}^{\!\top} \bm{A}\, \bm{S})}{\text{Tr}(\bm{S}^{\!\top} \bm{D}\, \bm{S}) + \epsilon}
\end{equation}
where $\bm{D} = \text{diag}(\bm{d})$ is the degree matrix. This formulation naturally allows imbalanced group sizes.

An assignment confidence loss encourages more confident community assignments:
\begin{equation}\label{eq:conf_loss}
    \mathcal{L}_{\text{conf}} = -\frac{1}{N}\sum_{i=1}^{N}\sum_{c=1}^{C} S_{ic} \log S_{ic}
\end{equation}

The combined community loss is $\mathcal{L}_{\text{comm}} = \mathcal{L}_{\text{cut}} + 0.5\, \mathcal{L}_{\text{conf}}$.



\subsubsection{Community Information Chunking}

When community information is available, the Global Interaction Scan~\eqref{eq:global_scan} re-orders agents by community ID before the Mamba scan, grouping same-community members consecutively. This allows the SSM to process entire communities as contiguous ``chunks,'' stabilizing gating behavior and enabling implicit segment-level processing. After the scan, agents are restored to their original ordering.

\begin{algorithm}[!htbp]
\caption{Community Feature Enrichment}
\label{alg:community}
\begin{algorithmic}[1]
\Require Node features $\bm{Z}\!\in\!\RR^{B\times N\times d}$; adjacency $\bm{A}\!\in\!\RR^{B\times N\times N}$; learnable prototypes $\bm{P}\!\in\!\RR^{C\times d}$; max communities $C$
\Ensure Enriched $\bm{Z}'$, community centrality $\bm{H}^{\text{comm}}$
\State \textit{\textbf{Differentiable community detection}}
\State $d_i\!\leftarrow\!\sum_j A_{ij}$ \Comment{the node degree}
\State Soft community assignments (mask $-\infty$ if $N\!<\!C$): 
\Statex $\bm{S}\!\leftarrow\!\softmax\!\bigl(\text{MLP}([\bm{Z};\bm{d}])\bigr)\!\in\!\RR^{B\times N\times C}$
\State \textit{\textbf{Feature enrichment}}
\State $\tilde{S}_{ic}\!\leftarrow\!S_{ic}/\sum_j S_{jc}$ \Comment{normalized assignment}
\State $\bm{c}_i^{\text{proto}}\!\leftarrow\!\sum_c S_{ic}\bm{P}_c$ \Comment{per-slot prototype context}
\State $\bm{c}_i^{\text{agg}}\!\leftarrow\!\sum_c \tilde{S}_{ic}(\bm{S}_{:c}^{\!\top}\bm{Z})$ \Comment{aggregated group context}
\State Residual fusion: 
\Statex $\bm{Z}'\!\leftarrow\!\text{LN}\!\bigl(\bm{Z}+\text{MLP}([\bm{Z};\,\bm{c}^{\text{proto}}\!+\!\bm{c}^{\text{agg}}])\bigr)$ 
\State $|\mathcal{C}_c|\!=\!\sum_j S_{jc}$ \Comment{soft community size}
\State Community-aware centrality~\eqref{eq:comm_centrality}:\Statex $H_i^{\text{comm}}\!\leftarrow\!\sum_c S_{ic}\,|\mathcal{C}_c|/N \;+\; 0.5\cdot \mathcal{H}(\bm{s}_i)/\log C$
\State return $\bm{Z}',\,\bm{H}^{\text{comm}}$
\end{algorithmic}
\end{algorithm}

\subsection{Dynamic Fusion and Global Scan}\label{sec:module5}

\subsubsection{Social Gate}

The three branches capture complementary perspectives with varying relevance depending on the scene context. We learn to fuse them with a dynamic gating mechanism:
\begin{align}
    \bm{w} &= \softmax\left(\text{MLP}\left([\bm{Z}^T_{:,-1}\;;\; \bm{Z}^E_{:,-1}\;;\; \bm{Z}^G_{:,-1}]\right)\right) \label{eq:gate_weight}\\
    \bm{Z}^{\text{fused}} &= w_T \odot \bm{Z}^T + w_E \odot \bm{Z}^E + w_G \odot \bm{Z}^G \label{eq:gate_fusion}
\end{align}
where $\bm{w}\in\RR^{3}$, the MLP operates on the last-timestep concatenation and produces per-agent, per-branch weights via softmax normalization. Since the ego and goal branches output sequences of length $T{+}1$, the prepended token at position~0 is discarded after the Mamba scan, retaining positions $1, \ldots, T$ which correspond to the original observation timesteps enriched with social context.

\subsubsection{Global Interaction Scan}

While the Social Triplet captures local pairwise interactions through ego/goal token injection, global population-level patterns (\eg, crowd flow direction, density gradients) require aggregation across all agents. We perform a final Mamba scan \emph{along the agent dimension}:
\begin{equation}\label{eq:global_scan}
    \bm{Z}_t^{\text{global}} = \text{MambaBlock}\left(\bm{Z}^{\text{fused}}[:, :, t, :]\right) \quad \forall t
\end{equation}

Implementation-wise, $\bm{Z}^{\text{fused}} \in \RR^{B \times N \times T \times d}$ is reshaped to $(BT, N, d)$ so each timestep is an independent batch element processed by the same MambaBlock along the $N$-agent sequence.

\subsection{Multimodal Trajectory Decoder and Training Objective}\label{sec:decoder}
\subsubsection{Decoder}
The decoder generates $K$ trajectory hypotheses from the final agent representation:
\begin{align}
    \bm{f}_i &= \text{MLP}_{\text{enc}}(\bm{Z}_i^{\text{global}}[:, -1, :]) \in \RR^{d} \label{eq:dec_encode}\\
    \hat{\bm{\Delta}}_i^{(k)} &= \text{MLP}_k(\bm{f}_i) \in \RR^{T_{\text{pred}} \times 2}, \quad k = 1, \ldots, K \label{eq:dec_traj}\\
    \hat{\bm{Y}}_i^{(k)} &= \text{cumsum}(\hat{\bm{\Delta}}_i^{(k)}) + \bm{p}_i^{T_{\text{obs}}} \label{eq:dec_anchor}\\
    \bm{\pi}_i &= \softmax\left(\text{MLP}_{\text{prob}}(\bm{f}_i)\right) \in \RR^{K} \label{eq:dec_mode}
\end{align}

Each of the $K$ modes has a dedicated MLP decoder~\eqref{eq:dec_traj}, ensuring architectural diversity among hypotheses. Mode probabilities~\eqref{eq:dec_mode} indicate the model's confidence in each hypothesis.

\subsubsection{Training Objective}\label{sec:training}

The total training loss combines multiple objectives:

Best-of-K Loss:
The primary trajectory loss selects the best hypothesis from $K$ predictions:
\begin{align}
    k^* &= \arg\min_k \left[\lambda_{\ade}\, \ade_k + \lambda_{\fde}\, \fde_k\right] \label{eq:best_k_select}\\
    \mathcal{L}_{\text{reg}} &= \lambda_{\ade}\, \ade_{k^*} + \lambda_{\fde}\, \fde_{k^*} \label{eq:reg_loss}\\
    \mathcal{L}_{\text{prob}} &= -\log \pi_{k^*} \label{eq:prob_loss}
\end{align}
where $\ade_k = \frac{1}{T_{\text{pred}}} \sum_t \|\hat{\bm{p}}_t^{(k)} - \bm{p}_t^{\text{gt}}\|_2$ and $\fde_k = \|\hat{\bm{p}}_{T_{\text{pred}}}^{(k)} - \bm{p}_{T_{\text{pred}}}^{\text{gt}}\|_2$.



Total Loss:
\begin{equation}\label{eq:total_loss}
    \mathcal{L} = \mathcal{L}_{\text{reg}} + \lambda_{\text{prob}}\, \mathcal{L}_{\text{prob}} + \lambda_{\text{comm}}\, \mathcal{L}_{\text{comm}}
\end{equation}
with $\lambda_{\text{prob}} = 0.1$, $\lambda_{\text{comm}} = 0.05$. The community loss $\mathcal{L}_{\text{comm}}$ is described in Sec.~\ref{sec:community}-2.

\subsubsection{Optimization}

We use AdamW~\cite{loshchilov2019decoupled} with learning rate $10^{-4}$, weight decay $10^{-4}$, and $(\beta_1, \beta_2) = (0.9, 0.999)$. The learning rate follows a cosine annealing schedule with linear warmup for 5 epochs and minimum learning rate $10^{-6}$. Gradient norms are clipped to 0.75 for SSM training stability. Mixed-precision training (FP16) is used for computational efficiency.

\begin{figure}[t]
    \centering
    \vspace{0.7mm}
    \centerline{\includegraphics[width=\linewidth]{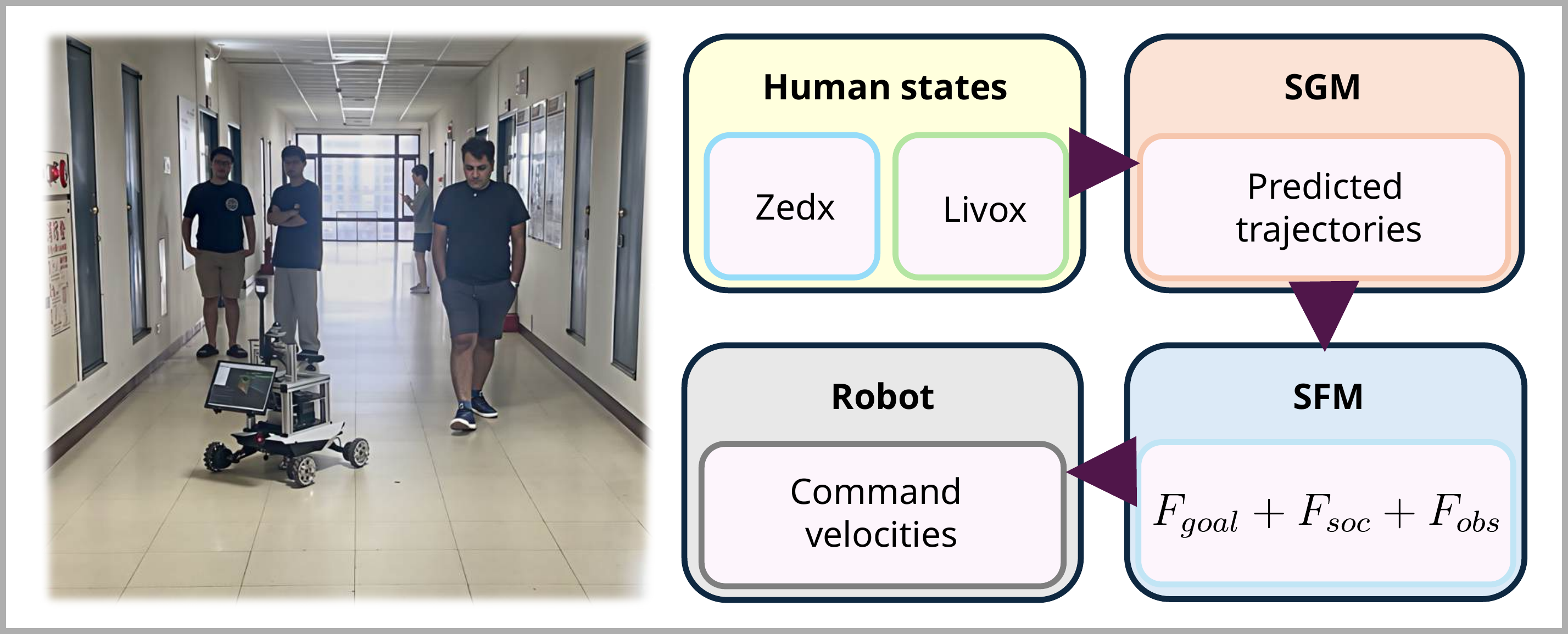}}
    \caption{Physical experiment setup}
    \label{fig:exp_setup}
\end{figure}

\section{Experiments}\label{sec:experiments}

\begin{table*}[t]
    \vspace{2mm}
    \centering
    \caption{Quantitative comparison on ETH/UCY benchmark (ADE/FDE in meters, minADE$_{20}$/minFDE$_{20}$). Best results in \textbf{bold}, second best \underline{underlined}. Results for comparison methods are from their original papers.}
    \label{tab:eth_ucy}
    \begin{tabular}{l|cc|cc|cc|cc|cc|cc}
        \toprule
        \multirow{2}{*}{Method} & \multicolumn{2}{c|}{ETH} & \multicolumn{2}{c|}{Hotel} & \multicolumn{2}{c|}{Univ} & \multicolumn{2}{c|}{Zara1} & \multicolumn{2}{c|}{Zara2} & \multicolumn{2}{c}{Average} \\
        & ADE & FDE & ADE & FDE & ADE & FDE & ADE & FDE & ADE & FDE & ADE & FDE \\
        \midrule
        Social-STGCNN~\cite{mohamed2020social} & 0.64 & 1.11 & 0.49 & 0.85 & 0.44 & 0.79 & 0.34 & 0.53 & 0.30 & 0.48 & 0.44 & 0.75 \\
        SGCN~\cite{shi2021sgcn} & 0.63 & 1.03 & 0.32 & 0.55 & 0.37 & 0.70 & 0.29 & 0.53 & 0.25 & 0.45 & 0.37 & 0.65 \\
        STAR~\cite{yu2020spatio} & \underline{0.36} & 0.65 & 0.17 & 0.36 & 0.31 & 0.62 & 0.26 & 0.55 & 0.22 & 0.46 & 0.26 & 0.53 \\
        PECNet~\cite{mangalam2020not} & 0.54 & 0.87 & 0.18 & 0.24 & 0.35 & 0.60 & 0.22 & 0.39 & 0.17 & 0.30 & 0.29 & 0.48 \\
        Trajectron++~\cite{salzmann2020trajectron++} & 0.43 & 0.86 & \textbf{0.12} & \textbf{0.19} & \textbf{0.22} & \textbf{0.43} & \textbf{0.17} & 0.32 & \textbf{0.12} & 0.25 & \textbf{0.21} & 0.41 \\
        AgentFormer~\cite{yuan2021agentformer} & 0.45 & 0.75 & 0.14 & 0.22 & \underline{0.25} & \underline{0.45} & \underline{0.18} & \underline{0.30} & 0.14 & \underline{0.24} & \underline{0.23} & 0.39 \\
        MART~\cite{lee2024mart} & \textbf{0.35} & \textbf{0.47} & 0.14 & 0.22 & \underline{0.25} & \underline{0.45} & \textbf{0.17} & \textbf{0.29} & \underline{0.13} & \textbf{0.22} & \textbf{0.21} & \textbf{0.33} \\
        \midrule
        \textbf{Ours} & 0.37 & \underline{0.52} & \underline{0.13} & \underline{0.21} & \underline{0.25} & 0.47 & 0.22 & 0.35 & 0.20 & 0.32 & \underline{0.23} & \underline{0.37} \\
        \bottomrule
    \end{tabular}
\end{table*}

\subsection{Datasets and Evaluation Metrics}

We evaluate SGM on two widely used pedestrian trajectory prediction benchmarks:

\textbf{ETH/UCY}~\cite{pellegrini2009you, lerner2007crowds} benchmark comprises five scenes: ETH, Hotel, Univ, Zara1, and Zara2. Following the standard leave-one-out protocol~\cite{alahi2016social}, we train on four scenes and test on the held-out scene. Each scene contains diverse crowd densities and interaction patterns. We use $T_{\text{obs}} = 8$ frames (3.2~s) to predict $T_{\text{pred}} = 12$ frames (4.8~s) at 2.5~FPS.

\textbf{Stanford Drone Dataset (SDD)}~\cite{robicquet2016learning} provides bird's-eye-view video from 8 unique scenes at a university campus with diverse agent types (pedestrians, bicyclists, skateboarders, cars). We follow the standard TrajNet split and predict 12 future frames from 8 observed frames.

\textbf{Simulated Crowd Dataset} (crowd\_sim~\cite {chen2019crowd} inspired simulation) is fast and cheap for controlled ablation studies; we generate synthetic robot-crowd interaction scenarios with configurable group structures using a SFM simulator. Each scenario contains one robot interacting with 15--29 pedestrians in 3--5 social groups of size 2--4. The simulation runs at 10~FPS with $T_{\text{obs}} = 10$ and $T_{\text{pred}} = 20$ frames.

\textbf{Evaluation Metrics}: Following standard protocols, we report 4 evaluation metrics. \textbf{ADE} (Average Displacement Error) is the average $\ell_2$ distance between predicted and ground-truth positions over all timesteps; \textbf{FDE} (Final Displacement Error) is the $\ell_2$ distance at the final predicted timestep; And \textbf{minADE$_K$} / \textbf{minFDE$_K$}: Minimum ADE/FDE among the best of $K$ trajectory hypotheses (oracle selection). We use $K = 20$.


\subsection{Implementation Details}

SGM uses hidden dimension $d = 256$, SSM state dimension $d_s = 16$, convolution dimension $d_c = 4$, and expansion factor 2 for Mamba. The Social Triplet uses 2 MambaBlock layers per branch. The decoder generates $K = 20$ trajectory hypotheses. Training uses batch size 64, 100 epochs, and the AdamW optimizer. All experiments are conducted on an NVIDIA RTX 4000 Ada GPU. The model has 1.55M parameters in the base configuration.

\begin{table}[t]
    \centering
    \caption{Quantitative comparison on SDD benchmark (minADE$_{20}$/minFDE$_{20}$ in pixels).}
    \label{tab:sdd}
    \begin{tabular}{l|cc}
        \toprule
        Method & minADE$_{20}$ & minFDE$_{20}$ \\
        \midrule
        Social-STGCNN~\cite{mohamed2020social} & 20.76 & 33.18 \\
        PECNet~\cite{mangalam2020not} & 9.96 & 15.88 \\
        Trajectron++~\cite{salzmann2020trajectron++} & 19.30 & 32.70 \\
        AgentFormer~\cite{yuan2021agentformer} & 8.05 & 12.45 \\
        MART~\cite{lee2024mart} & \underline{7.43} & \underline{11.82} \\
        \midrule
        \textbf{Ours} & \textbf{5.98} & \textbf{10.05} \\
        \bottomrule
    \end{tabular}
\end{table}

\subsection{Physical Experiment}
To validate the practical utility of SGM predictions, we integrate the trajectory forecasting module into a physical robot navigation system for social-aware behavior in pedestrian environments. The physical experiments are conducted in an indoor environment with 2-4 human participants walking in predefined group patterns. The physical setup represented in Fig.~\ref{fig:exp_setup}, we use an Agilex Scout Mini platform running ROS\,2 Humble. Pedestrian tracking is performed using the Zedx camera detection module. The SGM trajectory predictor provides multi-modal future trajectory estimates. The SFM controller follows the formulation proposed in~\cite{wu2022esfm} and reacts to collision avoidance based on predicted pedestrian trajectories. The SGM model runs inference on an onboard NVIDIA Jetson-Orin GPU at approximately 10~Hz. 

\begin{figure*}[t]
    \vspace{0.4mm}
    \centering
    \centerline{\includegraphics[width=\textwidth]{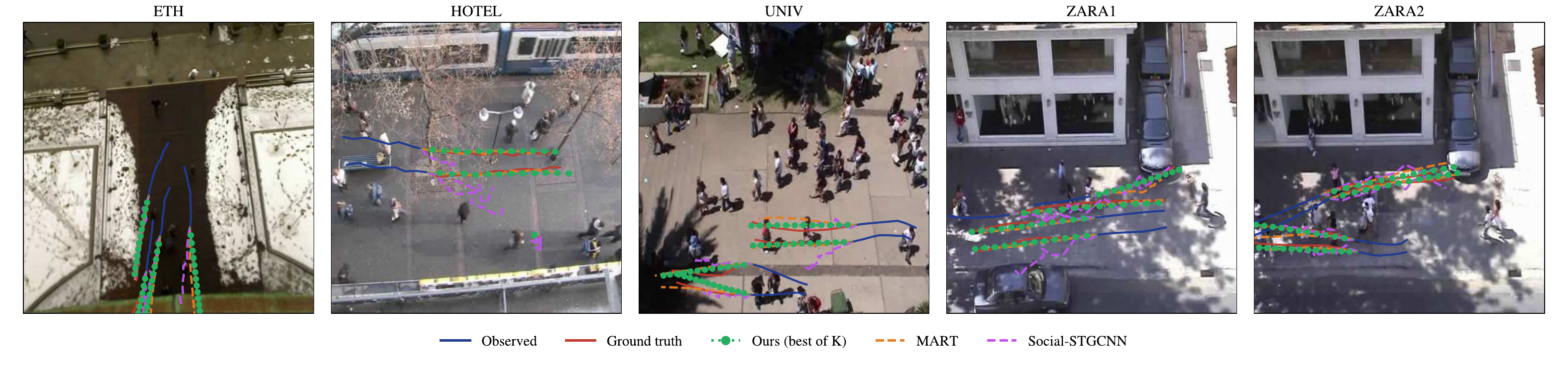}}
    \caption{Visualization of prediction on the ETH-UCY dataset.}
    \label{fig:viz_pubdataset_prediction}
\end{figure*}


\section{Results and Discussion}\label{sec:results_discussion}

\subsection{Quantitative Results}

We compare SGM against several state-of-the-art (SOTA) trajectory prediction methods, including graph convolutional approaches (Social-STGCNN~\cite{mohamed2020social} and SGCN~\cite{shi2021sgcn}), Transformer-based methods for social-temporal reasoning (STAR~\cite{yu2020spatio}, AgentFormer~\cite{yuan2021agentformer}), graph-based generative models (Trajectron++~\cite{salzmann2020trajectron++} and PECNet~\cite{mangalam2020not}), and MART~\cite{lee2024mart} which explicitly models group structures through interaction graphs. Table~\ref{tab:eth_ucy} summarizes the quantitative results on the ETH/UCY benchmark.

As shown in Table~\ref{tab:eth_ucy}, the proposed SGM achieves competitive performance across all benchmark scenes despite avoiding the use of computationally expensive self-attention mechanisms. In terms of overall performance, SGM attains the second-best average ADE/FDE of 0.23/0.37, outperforming most existing graph-based and Transformer-based baselines and remaining close to the best-performing method, MART.
On the other hand, Table~\ref{tab:sdd} presents the quantitative comparison on the SDD benchmark. Unlike the ETH/UCY benchmark, where SGM achieves competitive second-best performance, the proposed method establishes a new state-of-the-art on SDD, achieving the best minADE$_{20}$ and minFDE$_{20}$ scores of 5.98 and 10.05, respectively.

\subsection{Ablation Studies}

\subsubsection{Module ablation}
To understand the contribution of each component, we conduct comprehensive ablation studies on the simulated crowd dataset with controlled group structures. Table~\ref{tab:module_ablation} presents an ablation study evaluating the contribution of each proposed component. The full SGM model achieves the best performance across all evaluation metrics. Among all components, social triplet factorization contributes the most significantly to prediction performance. The Community-Aware module also provides consistent improvements. Without community-level reasoning, indicating that explicit modeling of group membership helps capture collective crowd behaviors beyond local interpersonal interactions. Finally, the minimal variant, which removes all proposed enhancements and keeps the dynamic interaction graph with the decoder only, exhibits the worst performance, with ADE and FDE increasing by 49.2\% and 36.1\%, respectively, compared to the full model. These results confirm that each module contributes positively to the overall framework.

\begin{table}[t]
    \centering
    \caption{Ablation study on the contribution of each SGM module. Evaluated on the simulated crowd dataset (ADE/FDE in meters).}
    \label{tab:module_ablation}
    \begin{tabular}{l|cccc}
        \toprule
        Configuration & minADE $\downarrow$ & minFDE $\downarrow$ & ADE $\downarrow$ & FDE $\downarrow$ \\
        \midrule
        SGM (Full)                   & 0.049 & 0.081 & 0.124 & 0.296 \\
        \midrule
        w/o Global scan              & 0.051 & 0.083 & 0.126 & 0.301 \\
        w/o Community-aware          & 0.051 & 0.085 & 0.129 & 0.310 \\
        w/o Social triplet           & 0.066 & 0.105 & 0.171 & 0.376 \\
        SGM (minimal)    & 0.070 & 0.116 & 0.185 & 0.403 \\
        \bottomrule
    \end{tabular}
\end{table}




\subsubsection{Number of Modes}

Table~\ref{tab:num_modes} indicates that while multi-modal prediction is essential for handling the stochastic nature of human behavior, the performance gain saturates beyond a moderate number of modes, highlighting a favorable trade-off between prediction quality and computational efficiency.

\begin{table}[t]
    \centering
    \caption{Effect of number of trajectory modes $K$.}
    \label{tab:num_modes}
    \begin{tabular}{c|cccc}
        \toprule
        $K$ & minADE $\downarrow$ & minFDE $\downarrow$ & ADE $\downarrow$ & FDE $\downarrow$  \\
        \midrule
        1   & 0.85 & 1.59 & 0.85 & 1.59 \\
        5   & 0.07 & 0.14 & 0.12 & 0.30 \\
        20  & 0.05 & 0.08 & 0.12 & 0.29 \\
        \bottomrule
    \end{tabular}
\end{table}

\subsubsection{Computational Efficiency}

We evaluate the computational efficiency of different sequence modeling backbones for social interaction modeling, as summarized in Table~\ref{tab:efficiency}, using a fixed setting of $N=50$ interacting agents. 
While all baseline methods rely on quadratic-complexity interaction modeling, the proposed SGM adopts a Selective State Space formulation with linear complexity $O(N)$. As shown in Table~\ref{tab:efficiency}, SGM achieves the lowest inference latency among all compared state-of-the-art methods, requiring only 4.27 ms per forward pass. This represents a 2.5$\times$ speedup over MART, the fastest Transformer-based baseline, and substantially outperforms other SOTA with attention-based approaches such as STAR and AgentFormer.


\begin{table}[t]
    \centering
    \caption{Computational comparison of sequence modeling backbones for social interaction modeling with $N=50$ agents.}
    \label{tab:efficiency}
    \begin{tabular}{l|ccc}
        \toprule
        Method & Backbone & Params (M) & Inference (ms) \\
        \midrule
        Social-STGCNN~\cite{mohamed2020social} & $O(N^2)$ & \textbf{0.04} & 121.39 \\
        STAR~\cite{yu2020spatio} & $O(N^2)$ & \underline{0.96} & 618.37 \\
        AgentFormer~\cite{yuan2021agentformer} & $O(N^2)$ & 6.50 & 13165.43 \\
        MART~\cite{lee2024mart} & $O(N^2)$ & 1.59 & \underline{10.72} \\
        \textbf{Ours} & \textbf{$O(N)$} & 1.55 & \textbf{4.27} \\
        \bottomrule
    \end{tabular}
\end{table}

\subsection{Qualitative Results}

Fig.~\ref{fig:viz_pubdataset_prediction} presents qualitative trajectory prediction (best-of-$K$ mode) results on the ETH/UCY benchmark, including the ETH, HOTEL, UNIV, ZARA1, and ZARA2 scenes from left to right. We compare the proposed SGM with two existing SOTA methods Social-STGCNN and MART to evaluate its ability to capture future motions. Across diverse crowd densities and interaction patterns, SGM consistently generates plausible future trajectories that closely follow the underlying pedestrian motion trends.

\begin{figure}[t]
\centering
\centerline{\includegraphics[width=\linewidth]{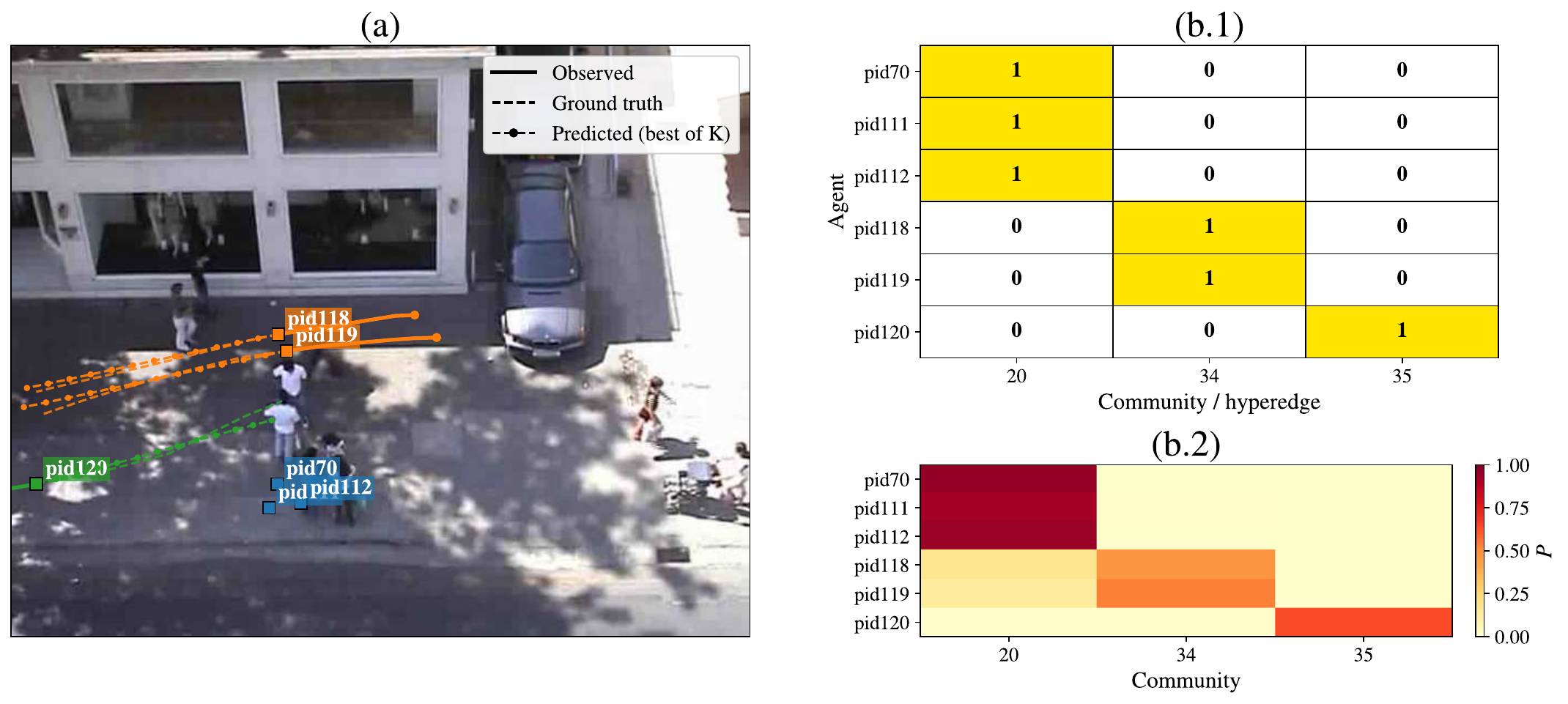}}
\caption{Community-aware representation}
\label{fig:comm_reasoning}
\end{figure}

To further validate the effectiveness of the proposed community-aware module, Fig.~\ref{fig:comm_reasoning}(a) provides a qualitative analysis of the learned group representations. Fig.~\ref{fig:comm_reasoning}(a) visualizes the unsupervised community discovery results, showing that the model can automatically identify coherent pedestrian groups without explicit group annotations. Furthermore, Fig.~\ref{fig:comm_reasoning}(b.1) illustrates the learned hyperedge incidence matrix, while Fig.~\ref{fig:comm_reasoning}(b.2) presents the corresponding heatmap representation.


\begin{table}[t]
    \centering
    \caption{Physical experiment results comparing baseline SFM controller (using relative positions) vs.~SGM-enhanced controller (using predicted trajectories) via ten trials.}
    \label{tab:physical}
    \begin{tabular}{l|cccc}
        \toprule
        Controller & Success & Time (s) & Clearance (m) & TTC \\
        \midrule
        Baseline SFM    & 66.7 & 30.56 & 0.65$\pm$0.21 & 0.84 \\
        SGM enhanced & 83.3 & 38.31 & 0.73$\pm$0.15 & 1.06 \\
        \bottomrule
    \end{tabular}
\end{table}

For the physical experiment, we measure: \emph{Success}--percentage of trials where the robot reaches its goal without collision; (2) \emph{Time}--average time to reach the destination; (3) \emph{Clearance}--closest distance to any pedestrian during navigation; and (4) \emph{Time-to-collision}--the minimum predicted time before a potential collision with a pedestrian. Table~\ref{tab:physical} evaluates the impact of integrating SGM predictions into the SFM during real-world robot navigation experiments via ten trials. Compared with the baseline SFM controller that relies solely on current pedestrian observations, the SGM-enhanced controller achieves a substantially higher navigation success rate, improving from 66.7\% to 83.3\%. The incorporation of predicted pedestrian trajectories also indicates that the robot maintains safer distances from surrounding pedestrians. Furthermore, the TTC metric suggests that the robot exhibits more socially compliant and anticipatory navigation behavior when future pedestrian motion is taken into account. Although the average navigation time gaps range from 30.56 s to 38.31 s, this behavior reflects a deliberate safety-efficiency trade-off. By anticipating future pedestrian movements rather than reacting only to instantaneous observations, the robot adopts more conservative avoidance maneuvers and proactively resolves potential conflicts before they become critical. Overall, these results demonstrate that the proposed trajectory prediction framework provides practical benefits beyond offline forecasting benchmarks. When integrated into a downstream navigation controller, SGM enables safer and more reliable robot navigation in dynamic human-populated environments while maintaining real-time operational capability. Additional experimental demonstrations are provided in the accompanying supplementary material.

\section{Conclusion}\label{sec:conclusion}

We presented SGM, a novel architecture for pedestrian trajectory prediction that replaces quadratic-complexity attention mechanisms with efficient SSMs. By constructing dynamic interaction graphs with physics-aware edge weights and ordering agents through topology-guided centrality, SGM introduces a principled framework for applying SSMs to unordered, graph-structured social data. Experimental results on the ETH/UCY and SDD benchmarks demonstrate that SGM achieves competitive prediction performance while avoiding the quadratic computational cost commonly associated with attention-based architectures. Beyond offline evaluation, we further validated the practical applicability of the proposed framework through physical robot experiments, where predicted pedestrian trajectories were integrated into the SFM controller for real-time navigation in dynamic environments. Future work includes extending SGM to heterogeneous agent types (pedestrians, cyclists, vehicles), incorporating scene context through occupancy maps, and deploying the full community-aware pipeline on real-world robot platforms.

\bibliographystyle{IEEEtran}
\bibliography{references}

\end{document}